\documentclass{article}

\usepackage{microtype}
\usepackage{graphicx}
\usepackage{subcaption}
\usepackage{booktabs} 

\usepackage{hyperref}

\usepackage[accepted]{icml2026}

\usepackage{amsmath}
\usepackage{amssymb}
\usepackage{mathtools}
\usepackage{amsthm}

\usepackage[capitalize,noabbrev]{cleveref}

\theoremstyle{plain}

\theoremstyle{definition}

\theoremstyle{remark}

\usepackage[textsize=tiny]{todonotes}

\usepackage{pifont} 
\usepackage{subcaption}  %
\usepackage{amsfonts}
\usepackage{wrapfig}
\usepackage{newfloat}
\usepackage{listings}
\usepackage{tikz}
\usepackage{tcolorbox}
\usepackage{multirow}
\tcbset{colback=white, colframe=black, boxrule=0.5pt}
\usepackage{xspace}

\newcommand{\papername}{ConsMSA\xspace}

\icmltitlerunning{ConsMSA: Semantic Distribution Consistency Learning for Multimodal Sentiment Analysis}

\begin{document}

\twocolumn[
  \icmltitle{ConsMSA: Semantic Distribution Consistency Learning for Multimodal Sentiment Analysis}



  \icmlsetsymbol{equal}{*}

  \begin{icmlauthorlist}
    \icmlauthor{Pan Wang}{1,2}
    \icmlauthor{Lipeng Ke}{2}
    \icmlauthor{Huajun Ying}{2}
    \icmlauthor{Pritish Mohapatra}{2}
    \icmlauthor{Rohan Sarkar}{2}
    \icmlauthor{Suresh Lakhani}{2}
    \icmlauthor{Sankar Venkataraman}{2}
    \icmlauthor{Jingtong Hu}{1}
  \end{icmlauthorlist}

  \icmlaffiliation{1}{Department of Electrical and Computer Engineering, University of Pittsburgh, Pittsburgh, USA}
  
  \icmlaffiliation{2}{Amazon, Sunnyvale, USA}

  \icmlcorrespondingauthor{Pan Wang}{pan.wang@pitt.edu}
  \icmlcorrespondingauthor{Jingtong Hu}{jthu@pitt.edu}


  \icmlkeywords{Multimodal Sentiment Analysis, Consistency Learning}

  \vskip 0.3in
]




\printAffiliationsAndNotice{}  


\begin{abstract}
Multimodal sentiment analysis (MSA) aims to predict human sentiments by integrating signals from different modalities such as text, video, and audio. However, raw multimodal sequences often suffer from semantic inconsistency--exhibiting redundancy or conflicts within and across modalities--which hinders robust understanding and increases computational cost. To this end, we introduce \papername, which explicitly formalizes semantic distribution consistency across both \textit{intra}- and \textit{inter}-modality, providing a consistency-aware mechanism for robust and efficient multimodal sentiment prediction. Specifically, \papername projects multimodal token features into a shared sentiment space to compute an Intra- and Inter-modality Consistency Score ($I^2CS$). By coupling this score with predictive relevance, we formulate consistency-aware importance signals that are utilized: (i) as a consistency regularizer to align latent distributions during training, (ii) to derive semantic-aware weights for adaptive multimodal token reweighting, and (iii) as a practical criterion to prune redundant or conflicting tokens.
Extensive experiments on CMU-MOSI, CMU-MOSEI, and CH-SIMS demonstrate that \papername achieves state-of-the-art performance while remaining robust under aggressive token compression--retaining only 10\% of tokens yields comparable accuracy. 
These results establish semantic distribution consistency as a promising foundation for synergizing predictive robustness with computational efficiency.
\end{abstract}

\section{Introduction}

Multimodal Sentiment Analysis (MSA) aims to infer human sentiments by integrating heterogeneous signals such as spoken language, facial expression, and vocal prosody~\citep{geetha2024multimodal,yang2022disentangled}. Modern MSA systems, especially Transformer-based, typically encode each modality into token-level representations and rely on cross-modal interaction mechanisms, such as cross-attention, to aggregate cues across time and modalities~\citep{wu2024multimodal}. Despite strong empirical progress, a persistent bottleneck remains: token-level evidence is often semantically inconsistent. This occurs \textit{intra}-modality due to pervasive redundancy or neutral signals, and \textit{inter}-modality due to semantic conflicts, such as those found in sarcasm. Such inconsistencies inject representational noise into the fusion process, forcing models to expend unnecessary computation on tokens that fail to support a coherent sentiment representation.

\begin{figure}[t]
  \centering
  \includegraphics[width=1\linewidth]{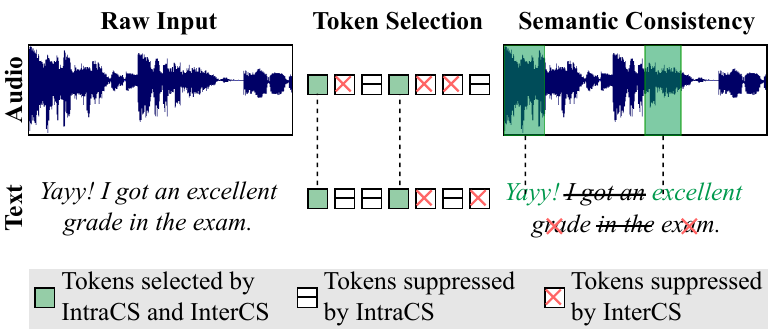}
  \caption{Motivation of ConsMSA. We explicitly model token-level \textit{intra-} and \textit{inter-}modality consistency by measuring \textbf{semantic alignment} in the prediction space, enabling multimodal \textbf{token reweighting and compression}.}
  \label{fig:BG}
\end{figure}

Existing work often addresses intra-modality structure learning and inter-modality interaction as largely separate objectives. Intra-modality modeling (e.g., self-attention) aims to preserve local structure and modality-specific semantics~\citep{tsai2018learning,hazarika2020misa}, whereas inter-modality modeling focuses on aligning cross-modal cues through cross-attention, contrastive objectives, or shared embedding spaces~\citep{yang2022disentangled,yang2022learning,li2023decoupled,yang2024clgsi,wu2024multimodal,wang2025dlf}. However, in MSA, the utility of token-level evidence ultimately depends on whether its semantic content is consistent with the latent sentiment outcome across modalities. When this semantic distribution consistency is not explicitly modeled, fusion modules can inadvertently overfit to redundant intra-modal tokens or amplify conflicting cross-modal noise.

As illustrated in Figure~\ref{fig:BG}, our approach is motivated by the principle of semantic distribution consistency, explicitly computing both intra-modality (IntraCS) and inter-modality (InterCS) consistency scores. However, while consistency effectively captures cross-modal agreement, it must be coupled with predictive relevance to avoid over-penalizing uncertain yet potentially informative signals.

To this end, we propose \papername, a consistency-aware framework that formalizes \textit{intra}- and \textit{inter}-modality consistency in the semantic distribution space via an Intra- and Inter-modality Consistency Score ($I^2CS$) based on Jensen–Shannon (JS) divergence~\citep{fuglede2004jensen}. Rather than using consistency in isolation, we operationalize practical signals that couple predictive relevance with $I^2CS$ to reward decision confidence while penalizing semantic-level disagreement. These consistency-aware importance signals serve as the core driver for:
(i) a consistency loss that enforces agreement on high-confidence and high-consistency tokens;
(ii) a differentiable gating mechanism that derives consistency-based weights for adaptive token reweighting;
(iii) a practical compression criterion to prune redundant or conflicting tokens while preserving the most informative evidence. By optimizing for this synergy between relevance and consensus, \papername yields a model that is both more accurate and remarkably compute-efficient.

Extensive experiments on CMU-MOSI~\citep{zadeh2016multimodal}, CMU-MOSEI~\citep{zadeh2018multimodal}, and CH-SIMS~\citep{yu2020ch} demonstrate that \papername not only achieves state-of-the-art performance but also exhibits remarkable robustness under aggressive token compression.

Our main contributions are summarized as follows:
\begin{itemize}
    \item \textbf{Holistic Consistency Paradigm.} We propose a novel framework, \papername,  which formalizes \textit{intra}- and \textit{inter}-modality semantic distribution consistency to address the pervasive challenges of multimodal noise and redundancy in MSA.
    \item \textbf{Information-Theoretic Consistency Metric}. In our framework, we introduce the $I^2CS$, a consistency-aware metric based on JS divergence to quantify semantic agreement within and across modalities in the semantic space.
    \item \textbf{Consistency-aware Importance Signals.} We formulate unified signals by coupling predictive relevance with $I^2CS$. These signals provide a criterion to reward tokens that are both highly informative and semantically coherent, guiding consistency-aware learning, differentiable feature reweighting, and practical token-level pruning.
    \item \textbf{Strong Performance.} \papername achieves state-of-the-art results on three widely used MSA benchmarks, CMU-MOSI~\citep{zadeh2016multimodal}, CMU-MOSEI~\citep{zadeh2018multimodal}, and CH-SIMS~\citep{yu2020ch}, and preserves competitive performance under substantial token reduction, demonstrating robustness and improved efficiency.
\end{itemize}

\section{Related Work}
\label{sec:related work}

\textbf{Disentangled and Aligned Representations for MSA.}
A major line of work in MSA improves fusion by disentangling modality-shared and modality-specific factors, thereby reducing distribution mismatch and encouraging complementary information. \citet{tsai2018learning} introduce a factorized formulation separating private and common components, and MISA~\citep{hazarika2020misa} extends this approach with explicit shared/private projections. Subsequent methods combine metric learning and adversarial objectives to align invariant and modality-specific representations \citep{yang2022disentangled,yang2022learning}, while DMD~\citep{li2023decoupled} addresses modality discrepancy via decoupled distillation. DLF~\citep{wang2025dlf} and MDF~\citep{jian2026mdf} further strengthen language-centric fusion via structured projections and targeted objectives.
These approaches primarily operate in feature space: they aim to learn better representations that can be fused, but they do not explicitly quantify whether token-level evidence is consistent with the downstream sentiment decision.

\textbf{Token Selection and Compression in Multimodal Learning.}
To reduce the cost of multimodal Transformers, prior work selects or prunes tokens using heuristics such as attention weights or entropy-based confidence~\citep{he2024zipvl,chen2024fastv,xing2025vision,wang2025models}, or by learning pruning modules that remove less informative tokens~\citep{tong2024flashsloth,zhang2024pmod}. While effective for acceleration, these methods typically score tokens in the token/feature space, which can be misaligned with the decision boundary of sentiment prediction and may remove tokens that are low-saliency but sentiment-critical.

\textbf{Our Perspective.}
In contrast, \papername formalizes multimodal consistency in the 
semantic distribution space. By measuring sentiment prediction-level agreement via JS divergence, and coupling it with relevance, we obtain a consistency-aware importance signal that simultaneously supports consistency regularization and token optimization (reweighting and compression). This bridges a key gap between representation learning and efficient inference in MSA.

\begin{figure*}[t]
  \centering
  \includegraphics[width=1\linewidth]{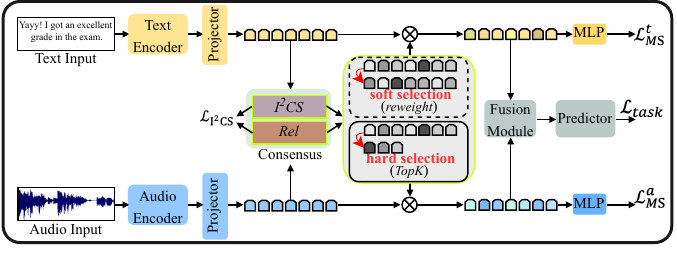}
  \caption{Overview of the ConsMSA framework. Modality-specific tokens are projected into a shared semantic space where the Consensus module computes \textit{intra}- and \textit{inter}-modality consistency ($I^2CS$). By coupling these scores with predictive relevance ($Rel$), we derive practical importance signals that drive consistency-aware learning and token optimization. The optimized tokens are finally concatenated in the Fusion Module for the final prediction. The system is optimized via a joint loss to achieve a robust balance between task accuracy ($\mathcal{L}_{task}$), semantic consistency ($\mathcal{L}_{I^2CS}$), and modality-specific losses ($\mathcal{L}_{MS}$).}
  \label{fig:framework}
\end{figure*}

\section{Method}

\subsection{Preliminaries}
Let $x^{(a)} \in \mathbb{R}^{L_a}$ and $x^{(t)} \in \mathbb{R}^{L_t}$ denote the raw audio waveform and the text sequence, respectively. The goal of MSA is to predict a sentiment label $y$ (categorical or continuous) from the multimodal input $(x^{(a)}, x^{(t)})$.
We encode the inputs into token-level features $h^{(a)} \in \mathbb{R}^{B \times T \times D}$ and $h^{(t)} \in \mathbb{R}^{B \times T \times D}$ (for simplicity, we omit modality superscripts and refer to tokens as $h$ throughout the rest of this paper), where $B$ is the batch size, $T$ is the token length, and $D$ is the feature dimension. Standard fusion architectures treat these features as the primary objects of interaction.
In contrast, \papername adopts a semantic distribution view: each token is projected to a latent sentiment prediction distribution, and multimodal learning is guided by how consistently these distributions support the same underlying sentiment.

\subsection{Overview}
As illustrated in Fig.~\ref{fig:framework}, \papername distills multimodal evidence through a three-stage pipeline. \textbf{First}, contextual tokens are projected into a shared semantic distribution space where the \textbf{Consensus} module computes intra- and inter-modality consistency ($I^2CS$) via JS divergence and predictive relevance ($Rel$). \textbf{Second}, we formulate practical importance signals by coupling $I^2CS$ with predictive relevance ($Rel$) to reward tokens that are both confident and mutually coherent. \textbf{Finally}, these signals are operationalized through three synergistic mechanisms: (i) a \textbf{consistency regularizer} (\textbf{$\mathcal{L}_{I^2\text{CS}}$}) for distribution alignment; (ii) a \textbf{soft gating mask} for adaptive feature reweighting; and (iii) a \textbf{hard selection operator} for practical token compression. The optimized features are concatenated in the Fusion Module, and then proceed to a Predictor for final prediction. The entire framework is optimized end-to-end to improve overall accuracy and computational efficiency.

\subsection{Intra- and Inter-modality Consistency Score($I^2CS$)}

In MSA, redundancy and conflict are two pervasive challenges that hinder robust representation learning. Redundancy occurs when many tokens within a modality induce near-identical (often neutral) sentiment evidence, diluting the contribution of the few sentiment-carrying tokens. Conflict arises when different modalities support incompatible interpretations (e.g., positive lexical content paired with a negative prosody), which can mislead fusion. We unify these phenomena under a single principle: token-level semantic inconsistency. 
Rather than crafting separate mechanisms for redundancy detection and cross-modal conflict resolution, we propose to directly measure and optimize intra- and inter-modality agreement in a space that is aligned with the downstream task--the sentiment prediction distribution space. This leads to an optimization objective that is both interpretable (agreement among predictions) and actionable (usable for weighting and pruning).

To this end, we define a consistency score using JS divergence.
Given two normalized distributions $p$ and $q$ (e.g., semantic-level predictions), the JS divergence is defined as:
\begin{equation}
\mathrm{JS}(p \| q) = \frac{1}{2} \mathrm{KL}(p \| m) + \frac{1}{2} \mathrm{KL}(q \| m), \quad m = \frac{1}{2}(p + q),
\end{equation}
where the Kullback-Leibler (KL) divergence is:
\begin{equation}
\mathrm{KL}(p(x) \| q(x)) = \sum_x p(x) \log \frac{p(x)}{q(x)}.
\end{equation}

In our formulation, $p$ and $q \in \mathbb{R}^{B \times L \times C}$ denote normalized prediction distributions, where $B$ is the batch size, $L$ is the token length, and $C$ is the number of semantic categories (e.g., sentiment classes). Throughout, we interpret each token as inducing a distribution over sentiment categories in a shared semantic space, so that divergence between distributions directly reflects semantic divergence.
Unlike KL divergence, Jensen-Shannon (JS) divergence is symmetric, finite, and strictly bounded, providing a highly stable mathematical foundation. Leveraging this, we formulate the $I^2CS$, a two-part objective that jointly models both intra- and inter-modal consistency.
\begin{itemize}
    \item \textbf{Intra-modality Consistency Score} measures within-modality semantic agreement by comparing the current token $h_i$ with all other tokens $h_j$ ($j \ne i$) from the same modality in the semantic distribution space:
    \begin{equation}
        \text{IntraCS}(h_i) = \frac{1}{T-1} \sum_{j \ne i} \text{JS}(p_i\| p_j),
        \label{eq:intracs}
    \end{equation}
    where $p_i$ and $p_j$ are the latent semantic distributions induced by $h_i$ and $h_j$ (e.g., via a semantic projector followed by \texttt{Softmax()}), and $T$ is the token length.

    \item \textbf{Inter-modality Consistency Score} measures cross-modality semantic agreement by comparing the current token $h_i$ in a source modality $p_i^{\text{src}}$ against tokens from other modalities $p_{j}^{\text{align}}$. Concretely, we average the JS divergence across all paired modalities and tokens:
    \begin{equation}
        \text{InterCS}(h_i) = \frac{1}{N} \sum_{n=1}^{N} \frac{1}{T} \sum^T_{j=1} \text{JS}(p_i^{\text{src}}\| p_{j}^{\text{align}}),
        \label{eq:intercs}
    \end{equation}
    where $p_i^{\text{src}}$ is the prediction distribution induced by the $i$th token of the source modality, and $p_{j}^{\text{align}}$ is the distribution induced by the corresponding token in a paired modality. $T$ is the token length, and $N$ is the number of modality pairs involved. For $M$ modalities, $N=M-1$ (e.g., for three modalities, $N=2$).
\end{itemize}

Thus, the overall $I^2CS$ for token $h_i$ is obtained by:
\begin{equation}
I^2CS(h_i) = \alpha \cdot \text{IntraCS}(h_i) + \beta \cdot \text{InterCS}(h_i),
\label{eq:i2cs}
\end{equation}
where $\alpha$ and $\beta$ balance intra-modality redundancy suppression and inter-modality conflict mitigation, respectively. By construction, a lower $I^2CS$ indicates higher consistency.

This score is the central signal in \papername: coupled with predictive relevance, it (i) regularizes training to encourage coherent token-induced predictions and (ii) drives token optimization (soft reweighting and hard pruning), yielding compact yet expressive multimodal representations.

\subsection{$I^2CS$-based Consistency Loss}

\begin{figure}[t]
  \centering
  \includegraphics[width=0.85\linewidth]{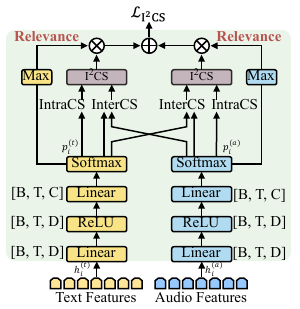}
  \caption{Details of our Consensus module. It projects text and audio tokens into a semantic space, and then computes IntraCS, InterCS, $I^2CS$, and relevance, resulting in $\mathcal{L}_{I^2CS}$.}
  \label{fig:I2C loss}
\end{figure}

To translate the consistency score into a trainable objective, we combine $I^2CS$ with a relevance term. Intuitively, we enforce consistency more strongly on tokens that are confident (i.e., likely to be sentiment-carrying), and we avoid over-penalizing uncertain tokens. As shown in Fig.~\ref{fig:I2C loss}, this yields the following relevance-weighted $I^2CS$ loss:
\begin{equation}
\begin{aligned}
    \mathcal{L}_{I^2CS} &= \frac{1}{T} \sum_{i=1}^{T} I^2CS(h_i) \cdot Rel(h_i) \\
    &= \frac{1}{T} \sum_{i=1}^{T} \Big[ \alpha \cdot \text{IntraCS}(h_i) \\
    &\quad \quad \quad \quad + \beta \cdot \text{InterCS}(h_i) \Big] \cdot Rel(h_i),
\end{aligned}
\label{eq:i2cs loss}
\end{equation}
where $Rel(h_i)$ measures token relevance, i.e., the confidence of the latent sentiment prediction:
\begin{equation}
    Rel(h_i) = \max(p_i),
\label{eq:rel}
\end{equation}
where $p_i$ is the latent semantic distribution, derived from:
\begin{equation}
\begin{aligned}
     p_i = Softmax(h_i^\prime), h_i^\prime = Proj2(ReLU(Proj1(h_i))),
\end{aligned}
\label{eq:pi}
\end{equation}
where $Proj1$ and $Proj2$ are the linear layer.

\subsection{Token Optimization Module via $I^2CS$}
The consistency score $I^2CS$ provides more than an auxiliary regularizer: it induces a decision-aligned ordering over tokens.
Recall that each token is mapped to a latent prediction distribution; thus, a token is useful for fusion if it provides reliable evidence (high confidence) and if this evidence is compatible with the consensus implied by other tokens within and across modalities (low divergence). This motivates a unified token-optimization principle: retain tokens whose predictions are both confident and mutually consistent.
We operationalize this principle by defining an importance signal that couples relevance and consistency:
\begin{equation}
    \text{Signal}(h_i) = Rel(h_i) - I^2CS(h_i).
    \label{eq:signal}
\end{equation}
The first term rewards decision confidence, whereas the second term penalizes semantic disagreement. Consequently, a large $\text{Signal}(h_i)$ identifies tokens that are not only sentiment-carrying but also coherent with the multimodal decision boundary, making them appropriate inputs to downstream fusion. Leveraging this signal, we derive two token optimization mechanisms: \textbf{Soft} and \textbf{Hard} selections.

\paragraph{Soft Selection.}
In the soft setting, we transform the importance signal into a differentiable weight:
\begin{equation}
    w_i = \sigma(\gamma \cdot \text{Signal}(h_i)), \quad w_i \in (0, 1),
    \label{eq:soft}
\end{equation}
where $\sigma$ is the sigmoid function and $\gamma$ controls the sharpness of the mapping. The soft mask $M_{\text{soft}} = [w_1, w_2, ..., w_T]$ then implements a 
consistency-aware gating
 mechanism: token representations are reweighted by $w_i$, so that evidence that is both confident and consistent contributes more strongly to downstream tasks.

\paragraph{Hard Selection.}
In the hard setting, we convert the soft weights into an explicit sparse subset using a \textit{TopK} operator:
\begin{equation}
    m_i = \textit{TopK}[w_i].
    \label{eq:hard}
\end{equation}
We retain only the top-$K$ tokens (largest $K$ $w_i$) for subsequent computations, yielding $M_{\text{hard}} = [m_1, m_2, ..., m_K]$. Importantly, this is not a heuristic pruning rule: it is the discrete counterpart of the same semantic-aligned criterion used in soft gating, and it therefore removes redundant or conflicting tokens while preserving the most informative, consistency-supported evidence.

As shown in Fig.~\ref{fig:framework}, the masks are derived from Eqs.~(\ref{eq:signal})-(\ref{eq:hard}), after which the optimized tokens are concatenated for the final prediction.

\begin{algorithm}[t]
\caption[End-to-end Pipeline]{\textbf{End-to-end \protect\papername\ Pipeline}}
\label{alg:i2c}
\textbf{Input:} Raw inputs $x^{(a)}, x^{(t)}$, sentiment label $y$ \\
\textbf{Output:} Prediction $\hat{y}$ and total loss $\mathcal{L}_{\text{joint}}$
\vspace{-2mm} 
\\
\noindent\rule{\linewidth}{0.2pt}

\textbf{\small Step 1: Feature Encoding} \par
{\centering $x^{(a)}, x^{(t)} \xrightarrow{\text{Encoders}} h^{(a)}, h^{(t)} \in \mathbb{R}^{B \times T \times D}$ \par}

\vspace{1mm}
\textbf{\small Step 2: Semantic Projection} \par
{\centering $h^{(a)}, h^{(t)} \xrightarrow{\text{Semantic Projector}} p^{(a)}, p^{(t)} \in \mathbb{R}^{B \times T \times C}$ \par}

\vspace{1mm}
\textbf{\small Step 3: $\text{I}^2\text{CS}$ Scoring} \par
{\centering $p^{(a)}, p^{(t)} \xrightarrow{\text{I}^2\text{CS Scorer}} \text{I}^2\text{CS}^{(a)}, \text{I}^2\text{CS}^{(t)} \in \mathbb{R}^{B \times T}$ \par}

\vspace{1mm}
\textbf{\small Step 4: Token Masking} \par
\vspace{-5mm}
\[
\vspace{-2mm}
\text{I}^2\text{CS}^{(a)}, \text{I}^2\text{CS}^{(t)} 
\xrightarrow[\text{Rel}]{\text{Masking}}
\begin{cases}
\tilde{h}^{(a)}, \tilde{h}^{(t)} \in \mathbb{R}^{B \times T \times D}, \text{(soft)} \\
\tilde{h}^{(a)}, \tilde{h}^{(t)} \in \mathbb{R}^{B \times K \times D}, \text{(hard)}
\end{cases}
\]

\vspace{1mm}
\textbf{\small Step 5: Prediction and Modality Heads} \par
{\centering $\begin{aligned}
\tilde{h}^{(a)}, \tilde{h}^{(t)} &\xrightarrow{\text{Fusion}} \hat{y} \quad \quad \quad \quad \\
\tilde{h}^{(a)} \xrightarrow{\text{MLP}_a} \hat{y}_a, &\quad
\tilde{h}^{(t)} \xrightarrow{\text{MLP}_t} \hat{y}_t
\end{aligned}$ \par}

\vspace{1mm}
\textbf{\small Step 6: Train using Joint Loss} \par
{\centering $\mathcal{L}_{\text{joint}} = \lambda_1 \mathcal{L}_{\text{task}} + \lambda_2 \mathcal{L}_{I^2CS} + \lambda_3 \mathcal{L}_{\text{MS}}$ \par}

\vspace{1mm}
\end{algorithm}

\subsection{Final Learning Objective}

\paragraph{Task Loss.}

We supervise the multimodal prediction with a standard $\ell_1$ regression loss. Given a batch of predicted sentiment scores $\hat{y} \in \mathbb{R}^N$ and ground-truth labels $y \in \mathbb{R}^N$, the task loss is:
\begin{equation}
    \mathcal{L}_{\text{task}} = \frac{1}{N} \sum_{i=1}^{N} \left| \hat{y}_i - y_i \right|.
    \label{eq:task loss}
\end{equation}
Here $N$ denotes the number of samples in the batch. This loss directly optimizes decision accuracy in the target space and serves as the primary supervision signal.

\paragraph{Modality-specific Losses (MS losses).}

In addition to the multimodal task loss, we impose modality-specific supervision before fusion. This encourages each modality to remain individually predictive and prevents the fusion module from relying on spurious cross-modal correlations. The modality-specific loss is:
\begin{equation}
    \mathcal{L}_{\text{MS}} = \sum_{m \in \{t, a\}} \lambda_m \cdot l_m = \sum_{m \in \{t, a\}} \lambda_m \cdot | \hat{y}_m - y_{\text{true}} |,
    \label{eq:ms loss}
\end{equation}
where $\hat{y}_m$ is the unimodal prediction from modality $m$ and $y_{\text{true}}$ is the ground-truth label.

\paragraph{Joint Learning Objective.}
Overall, we train the model by jointly optimizing prediction accuracy, consistency regularization, and modality-specific supervision:
\begin{equation}
    \mathcal{L}_{\text{joint}} = \lambda_1 \mathcal{L}_{\text{task}} + \lambda_2 \mathcal{L}_{I^2CS} + \lambda_3 \mathcal{L}_{\text{MS}}.
    \label{eq:total loss}
\end{equation}
The coefficients $\lambda_1$, $\lambda_2$, and $\lambda_3$ balance task supervision with the proposed semantic distribution consistency objective and unimodal auxiliary training. Finally, we summarize the end-to-end pipeline of ConsMSA in Algorithm~\ref{alg:i2c}.

\section{Experiments}
\label{sec:experiments}

\subsection{Datasets and Evaluation Metrics}
We evaluate \papername on three widely used datasets: \textbf{CMU-MOSI}~\citep{zadeh2016multimodal}, \textbf{CMU-MOSEI}~\citep{zadeh2018multimodal}, and \textbf{CH-SIMS}~\citep{yu2020ch}, all containing annotated video clips with aligned audio, visual, and textual information.

\textbf{CMU-MOSI.} This dataset consists of 2,199 short video segments, where each segment features a speaker expressing an opinion in monologue form. The data split includes 1,284 training instances, 229 for validation, and 686 for testing.

\textbf{CMU-MOSEI.} As a larger-scale benchmark, MOSEI contains 22,856 video clips collected from online movie reviews. Similar to MOSI, each clip is annotated with fine-grained sentiment labels on a continuous scale from $-3$ (strongly negative) to $+3$ (strongly positive). The dataset adopts a commonly used partitioning: 16,326 samples for training, 1,871 for validation, and 4,659 for testing.

\textbf{CH-SIMS.} As a widely used Chinese MSA benchmark, CH-SIMS consists of 2,281 video segments collected from various movies, television series, and variety shows. Similar to MOSEI, each segment is annotated with fine-grained sentiment labels, but on a continuous scale ranging from $-1$ (strongly negative) to $+1$ (strongly positive). The dataset follows a standard partitioning, comprising 1,368 samples for training, 456 for validation, and 457 for testing.

\begin{table*}[t]
\centering
\caption{Comparison of our method with prior SOTA methods on MOSI and MOSEI datasets. 
The best performance is highlighted in bold. 
T: Text, A: Audio, and V: Video. 
Note: $^{\dagger}$ denotes the result from THUIAR's GitHub page, $^*$ denotes the result from \citep{hazarika2020misa}, 
- denotes that the result from the original paper is not provided, and others are all from the original papers. For fair comparison, the BERT-based models utilize T+A+V following DLF, while the RoBERTa-based model utilizes T+A following MMML.}
\small
\begin{tabular}{l c cccc cccc}
\toprule
\multirow{2}{*}{Method} & \multirow{2}{*}{Modality} & \multicolumn{4}{c}{CMU-MOSI} & \multicolumn{4}{c}{CMU-MOSEI} \\
\cmidrule(lr){3-6} \cmidrule(lr){7-10}
& & Acc-7($\uparrow$) & Acc-2($\uparrow$) & F1($\uparrow$) & MAE($\downarrow$) & Acc-7($\uparrow$) & Acc-2($\uparrow$) & F1($\uparrow$) & MAE($\downarrow$) \\
\midrule
\multicolumn{10}{c}{BERT-based} \\
\midrule
TFN*                 & T+A+V & 34.90 & 80.08 & 80.07 & 0.901 & 50.20 & 82.50 & 82.10 & 0.593 \\
LMF*                 & T+A+V & 33.20 & 82.50 & 82.40 & 0.917 & 48.00 & 82.00 & 82.10 & 0.623 \\
EF-LSTM$^{\dagger}$  & T+A+V & 35.39 & 78.48 & 78.51 & 0.949 & 50.01 & 80.79 & 80.67 & 0.601 \\
LF-DNN$^{\dagger}$   & T+A+V & 34.52 & 78.63 & 78.63 & 0.955 & 50.83 & 82.74 & 82.52 & 0.580 \\
MFN$^{\dagger}$      & T+A+V & 35.83 & 78.87 & 78.90 & 0.927 & 51.34 & 82.85 & 82.85 & 0.575 \\
Graph-MFN$^{\dagger}$& T+A+V & 34.64 & 78.35 & 78.35 & 0.956 & 51.37 & 83.48 & 83.43 & 0.575 \\
MulT                 & T+A+V & 40.00 & 83.00 & 82.00 & 0.871 & 51.80 & 82.50 & 82.30 & 0.580 \\
PMR                  & T+A+V & 40.60 & 83.60 & 83.60 & -     & 52.50 & 83.60 & 83.40 & -     \\
MISA$^{\dagger}$     & T+A+V & 41.37 & 83.54 & 83.58 & 0.777 & 52.05 & 84.67 & 84.66 & 0.558 \\
MAG-BERT             & T+A+V & 43.62 & 84.43 & 84.61 & 0.727 & 52.67 & 84.82 & 84.71 & 0.543 \\
FDMER                & T+A+V & 44.10 & 84.60 & 84.70 & 0.724 & 54.10 & 86.10 & 85.80 & 0.536 \\
DMD                  & T+A+V & 45.60 & 86.00 & 86.00 & -     & 54.50 & 86.60 & 86.60 & -     \\
MMIM                 & T+A+V & 46.65 & 86.06 & 85.98 & 0.700 & 54.24 & 85.97 & 85.94 & 0.526 \\
Self-MM$^{\dagger}$  & T+A+V & 46.67 & 85.98 & 85.95 & 0.713 & 53.87 & 85.17 & 85.30 & 0.530 \\
DLF                  & T+A+V & 47.08 & 85.06 & 85.04 & 0.731 & 53.90 & 85.42 & 85.27 & 0.536 \\
ULMD                 & T+A+V & 47.81 & 85.82 & 85.71 & 0.700 & 53.81 & 85.95 & 85.91 & 0.531 \\
EMOE                 & T+A+V & 47.70 & 85.40 & 85.40 & 0.710 & 54.10 & 85.30 & 85.30 & 0.536 \\
MDF                  & T+A+V & \textbf{48.25} & 85.82 & 85.90 & \textbf{0.692} & 54.80 & 86.61 & 86.41 & 0.525 \\
\textbf{\papername (Ours)} & T+A+V & 47.96 & \textbf{86.06} & \textbf{86.06} & 0.698 & \textbf{54.92} & \textbf{86.67} & \textbf{86.65} & \textbf{0.525 }\\
\midrule
\multicolumn{10}{c}{RoBERTa-based} \\
\midrule
MMML                 & T+A   & 50.34 & 89.69 & 89.67 & 0.580 & 55.74 & 88.02 & 88.15 & 0.492 \\
\textbf{\papername (Ours)} & T+A   & \textbf{51.46} & \textbf{90.09} & \textbf{90.08} & \textbf{0.557} & \textbf{56.00} & \textbf{88.41} & \textbf{88.39} & \textbf{0.485} \\
\bottomrule
\end{tabular}
\label{tab:comparison}
\end{table*}

\textbf{Evaluation Metrics.} Following standard evaluation protocols in prior studies~\citep{liang2021attention,lv2021progressive,mao2022m}, we assess model performance using several metrics. These include 7-class accuracy (Acc-7), binary accuracy (Acc-2), F1 score, and mean absolute error (MAE). (For CH-SIMS, Acc-5, Acc-3, Acc-2, F1, and MAE are used.) These metrics provide a comprehensive view of the model's effectiveness for sentiment analysis.

\subsection{Implementation Details} 

We implement all experiments using PyTorch on a single NVIDIA A100 GPU with 40GB of memory. For the text modality, we adopt RoBERTa \citep{liu2019roberta} as the encoder, generating 1024-dimensional word-level embeddings. For the audio modality, we employ Data2Vec-Audio \citep{baevski2022data2vec}, which produces 768-dimensional token representations. The \papername model is trained using the AdamW optimizer with an initial learning rate of $5 \times 10^{-6}$ and a batch size of 8. Following the training setup in \citep{wu2024multimodal}, we train the model until convergence, using early stopping with a patience of 8 epochs. Additionally, we utilize BERT \citep{devlin2019bert} as the text encoder to compare with previous SOTA models like MDF~\citep{jian2026mdf}. See Appendix for more details.

\subsection{Main Results}

\textbf{Baselines.} To assess the effectiveness of our method, we compare \papername against a comprehensive set of competitive baselines on both datasets. The selected methods span a broad spectrum of representative MSA strategies, including early fusion (\textbf{EF-LSTM} \citep{williams2018recognizing}, \textbf{LF-DNN} \citep{williams2018dnn}), memory-augmented frameworks (\textbf{MFN} \citep{zadeh2018memory}, \textbf{Graph-MFN} \citep{zadeh2018multimodal}), attention-based and transformer-based approaches (\textbf{MulT} \citep{tsai2019multimodal}, \textbf{TFN} \citep{zadeh2017tensor}, \textbf{LMF} \citep{liu2018efficient}, \textbf{MAG-BERT} \citep{rahman2020integrating}, and \textbf{PMR} \citep{lv2021progressive}), multi-task learning (\textbf{Self-MM} \citep{yu2021learning}), as well as more recent decoupled or multi-loss techniques (\textbf{MISA} \citep{hazarika2020misa}, \textbf{FDMER} \citep{yang2022disentangled}, \textbf{DMD} \citep{li2023decoupled}, \textbf{MMIM} \citep{han2021improving}, \textbf{MMML} \citep{wu2024multimodal}, \textbf{DLF} \citep{wang2025dlf}, \textbf{ULMD}~\citep{ZHU2025102787}, \textbf{EMOE}~\citep{fang2025emoe}, and \textbf{MDF}~\citep{jian2026mdf}). More details are in Appendix.

\paragraph{Performance Comparison.}
Table~\ref{tab:comparison} shows that \papername consistently outperforms a broad range of prior MSA methods on both benchmarks. Relative to earlier fusion-based baselines, the gains are substantial, indicating that explicitly modeling semantic consistency provides stronger supervision than direct feature concatenation and generic cross-modal interaction alone. More importantly, \papername outperforms recent SOTA systems and achieves the best results across all metrics in both the BERT-based (T+A+V) and RoBERTa-based (T+A) settings.
This cross-backbone improvement suggests that our contribution is not tied to a particular encoder; rather, it stems from the proposed consistency-aware scoring and optimization, which helps the model focus on semantically aligned, sentiment-relevant tokens while suppressing redundant or conflicting evidence. More results are reported in the Appendix.

\paragraph{Discussion.}
The results provide three key takeaways.
\textit{First}, semantic consistency is a reliable learning signal for MSA: optimizing intra- and inter-modality agreement in the semantic distribution space (via $I^2CS$) encourages representations that are both discriminative and coherent, which benefits classification metrics (Acc-7/Acc-2/F1) and reduces regression error (MAE).
\textit{Second}, the gains are complementary to stronger unimodal encoders: replacing BERT with RoBERTa improves the textual backbone, yet \papername still delivers additional gains, indicating that the proposed token-level scoring, reweighting, and compression are additive rather than redundant.
\textit{Third}, efficiency and robustness follow naturally from the same principle: because $I^2CS$ highlights tokens that are both confident (high relevance) and semantically aligned (low divergence), the model can retain a small subset of informative tokens without sacrificing performance, which is consistent with the strong results under heavy token pruning.
Overall, \papername offers a practical mechanism to simultaneously improve effectiveness and efficiency by explicitly targeting the core failure mode in MSA---semantic inconsistency.

\textbf{Comparison on the CH-SIMS Dataset.} 
To validate the generalization of our method in a different linguistic context, we evaluate its performance on the Chinese MSA dataset, CH-SIMS. Currently, MMML stands as the SOTA on this benchmark. To ensure a strictly fair and rigorous comparison, we report our results solely under the bi-modal (``T+A'') setting in the main text, thereby precisely aligning with the established evaluation protocol of MMML. The comprehensive results for the complete tri-modal (``T+A+V'') configuration are deferred to the Appendix. As illustrated in Table~\ref{tab:ch_sims_comparison}, our approach consistently surpasses the strong MMML baseline across all evaluation metrics. Most notably, our model yields substantial absolute improvements of 1.31\% in Acc-3 (from 69.37\% to 70.68\%) and 1.39\% in Acc-5 (from 49.38\% to 50.77\%), while also achieving a higher F1 score (83.64\%). These consistent empirical gains underscore the efficacy of our proposed architecture: our method is capable of extracting highly discriminative features and achieving superior cross-modal alignment, successfully establishing a new SOTA under this setting.

\begin{table}[t]
\centering
\small
\caption{Comparison on CH-SIMS dataset.}
\label{tab:ch_sims_comparison}
\setlength{\tabcolsep}{0.8mm}
\begin{tabular}{llccccc}
\toprule
\textbf{Method} & \textbf{Modality} & \textbf{Acc-2}~$\uparrow$ & \textbf{Acc-3}~$\uparrow$ & \textbf{Acc-5}~$\uparrow$ & \textbf{F1}~$\uparrow$ & \textbf{MAE}~$\downarrow$ \\
\midrule
MMML  & T+A   & 82.93 & 69.37 & 49.38 & 82.90 & 0.332 \\
\midrule
\textbf{Ours}   & \textbf{T+A}   & \textbf{83.59} & \textbf{70.68} & \textbf{50.77} & \textbf{83.64} & \textbf{0.331} \\
\bottomrule
\end{tabular}
\end{table}

\subsection{Ablation Study}
\label{sec:ab}

Ablation studies on MOSI aim to verify the contribution of each component in \papername. Specifically, we evaluate (i) the two consistency terms (IntraCS and InterCS), (ii) the relevance estimation used in the loss and masking signal ($Rel$ in Eq.~(\ref{eq:i2cs loss}) and Eq.~(\ref{eq:signal})), (iii) the soft gating mechanism, and (iv) the two auxiliary objectives for unimodal supervision and consistency regularization 
($\mathcal{L}_{MS}$ and $\mathcal{L}_{I^2CS}$).

\textbf{Different Components.}
Table~\ref{tab:A2} summarizes component-wise ablations on MOSI. The results support the design choices of \papername from three perspectives.
\textit{Consistency modeling:} removing either IntraCS or InterCS causes clear degradation, confirming that both within-modality redundancy and cross-modality conflict matter in practice, and that addressing only one of them is insufficient.
\textit{Faithful token selection:} removing the relevance terms $Rel$ (Eq.~(\ref{eq:i2cs loss}) and Eq.~(\ref{eq:signal})) also hurts performance, suggesting that consistency should be interpreted together with confidence; penalizing or pruning low-consistency tokens without accounting for relevance can discard sentiment-carrying evidence.
\textit{Training and optimization synergy:} disabling the gating mechanism (Eq.~(\ref{eq:soft})) or removing either auxiliary loss ($\mathcal{L}_{MS}$ or $\mathcal{L}_{I^2CS}$) consistently decreases performance. This indicates that (a) token optimization provides a better feature set for fusion, and (b) explicit unimodal supervision and consistency regularization stabilize learning, yielding a more robust multimodal predictor.
Overall, the ablation study shows that performance gains of \papername are not attributable to a single component; they arise from the coherent integration of consistency scoring, relevance-aware token optimization, and joint supervision.

\begin{table}[t]
\small
\centering
\caption{Results of ablation studies on the MOSI benchmark.}
\setlength{\tabcolsep}{1.0mm}
\begin{tabular}{lcccc}
\toprule
Method & Acc-7($\uparrow$) & Acc-2($\uparrow$) & F1($\uparrow$)  & MAE($\downarrow$) \\
\midrule
\textbf{\papername (Ours)} & \textbf{51.46} & \textbf{90.09} & \textbf{90.08} & \textbf{0.557} \\
\midrule
w/o IntraCS & 47.52 & 87.20 & 87.21 & 0.629 \\
w/o InterCS & 46.79 & 88.87 & 88.83 & 0.626 \\
w/o $Rel$ in Eq.~(\ref{eq:i2cs loss}) & 45.48 & 87.96 & 87.99 & 0.668  \\
w/o $Rel$ in Eq.~(\ref{eq:signal}) & 46.36 & 87.80 & 87.82 & 0.647 \\
w/o Gating & 45.77 & 87.20 & 87.21 & 0.653 \\
w/o $\mathcal{L}_{MS}$ & 47.38 & 87.50 & 87.51 & 0.643 \\
w/o $\mathcal{L}_{I^2CS}$ & 48.83 & 86.59 & 86.62 & 0.634 \\
\bottomrule
\end{tabular}
\label{tab:A2}
\end{table}

\subsection{Further Analysis}

\textbf{Token Compression Robustness and Generalization.} 
To investigate the performance dynamics under different token retention regimes, we sweep the retention ratio from 1.0 down to 0.1 across both the CH-SIMS and CMU-MOSI datasets, as summarized in Table~\ref{tab:ablation_retention}. A ratio of 1.0 denotes the soft selection setting, where the entire sequence is retained but reweighted by the importance signal. Conversely, ratios below 1.0 represent the hard selection setting, where a Top-$K$ operator explicitly prunes low-consistency tokens. The empirical results reveal a compelling finding: on the MOSI dataset, hard selection at a ratio of 0.8 (Acc-2 90.85\%) outperforms soft selection at 1.0 (Acc-2 90.09\%), providing strong validation for our framework. Rather than indicating a limitation, this performance gain demonstrates that $\text{I}^2\text{CS}$-driven pruning acts as an effective ``semantic purifier''. While soft selection downweights inconsistent evidence, explicitly removing ``semantic noise''---tokens whose induced predictions conflict with multimodal consensus---further sharpens the representative quality of the features prior to fusion. Notably, while the optimal retention ratio varies across datasets (0.6 for CH-SIMS vs. 0.8 for MOSI), likely due to inherent differences in linguistic data redundancy (e.g., Chinese vs. English semantics), the core conclusion remains consistent: our method consistently matches or exceeds full-sequence performance via strategic token retention. Furthermore, the fact that performance remains highly robust even under extreme compression (retaining only 10\% of tokens at a 0.1 ratio) underscores that our consistency learning accurately isolates the sparse, sentiment-carrying core of the sequence, effectively distinguishing it from redundant or conflicting signals across diverse MSA tasks.

\begin{table}[t]
\small
\centering
\caption{Different token retention ratios on CH-SIMS and MOSI.}
\label{tab:ablation_retention}
\setlength{\tabcolsep}{1mm}
\begin{tabular}{ccc}
\toprule
\textbf{Ratio} & \textbf{CH-SIMS (Acc2/F1/MAE)} & \textbf{MOSI (Acc2/F1/MAE)} \\
\midrule
\textbf{1.0} & 83.59/83.64/\textbf{0.331} & 90.09/90.08/0.557 \\
\textbf{0.8} & 82.06/82.25/0.341          & \textbf{90.85/90.84/0.553} \\
0.7          & 83.15/83.20/0.347          & 89.33/89.34/0.562 \\
\textbf{0.6} & \textbf{83.59/83.70}/0.337 & 89.63/89.66/0.585 \\
0.5          & 82.71/83.00/0.337          & 88.72/88.75/0.585 \\
0.3          & 81.84/82.19/0.333          & 89.33/89.31/0.567 \\
0.1          & 81.18/81.32/0.355          & 89.18/89.17/0.576 \\
\bottomrule
\end{tabular}
\end{table}

\begin{figure*}[ht]
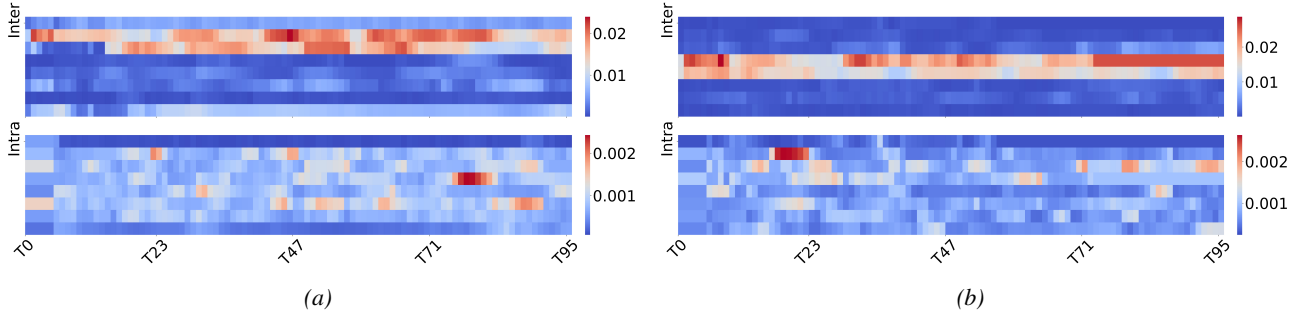

    \centering
    \subfloat[]{
        \includegraphics[width=0.488\textwidth]{./Figs/pre.pdf}
        \label{fig:first_v2}
    }
    \subfloat[]{
        \includegraphics[width=0.488\textwidth]{./Figs/after.pdf}
        \label{fig:second_v2}
    }
    \caption{Visualization of intra- and inter-modality semantic consistency at the token level. (a) Pre-optimization with \papername framework. (b) After-optimization with \papername framework. \textit{Note}: Lower value, higher consistency.}
    \label{fig:token_heatmap}
\end{figure*}

\begin{figure}[ht]
\centering
\subfloat[]{
    \includegraphics[width=0.45\linewidth]{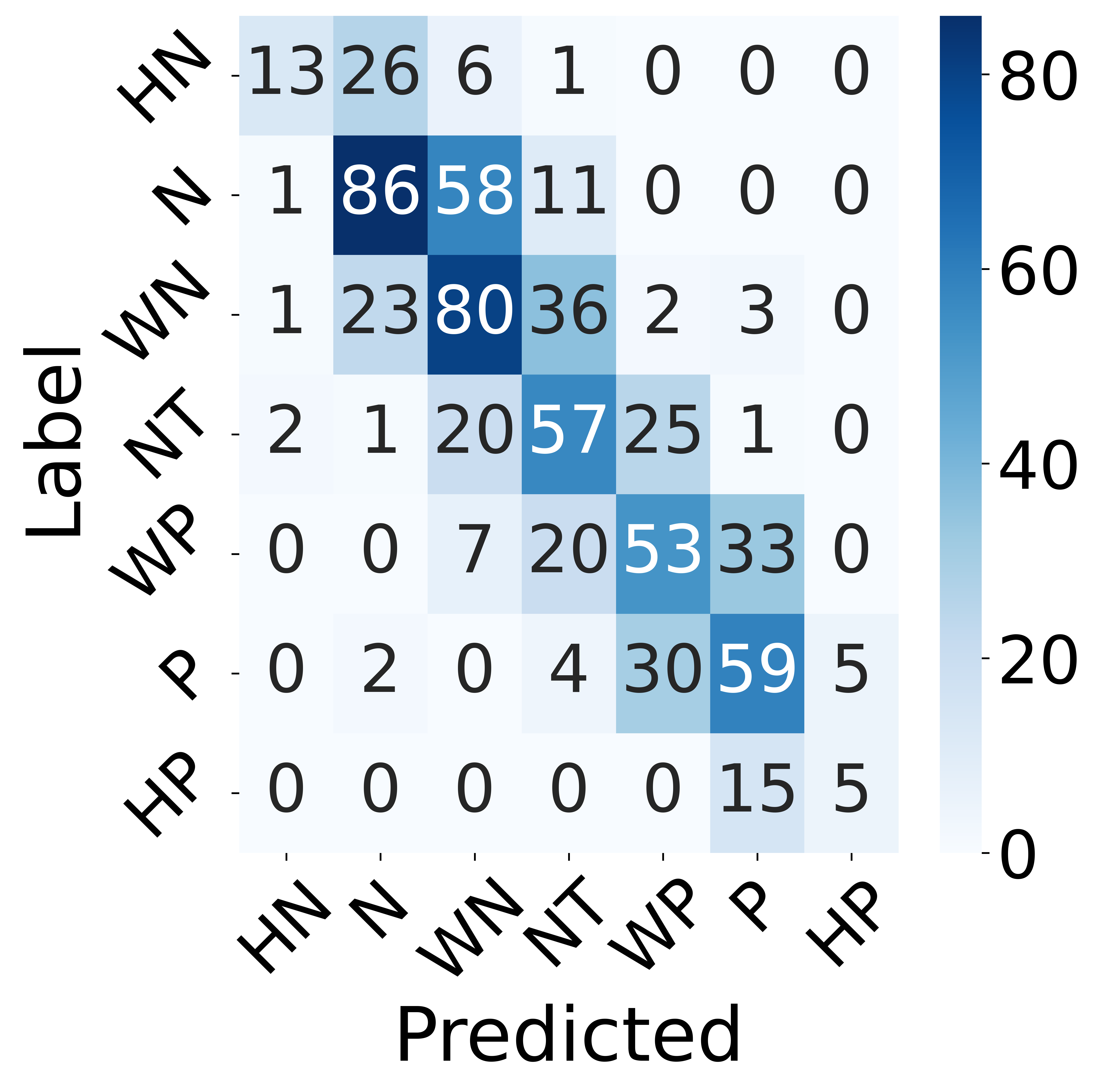}
    \label{fig:first}
}
\;
\subfloat[]{
    \includegraphics[width=0.45\linewidth]{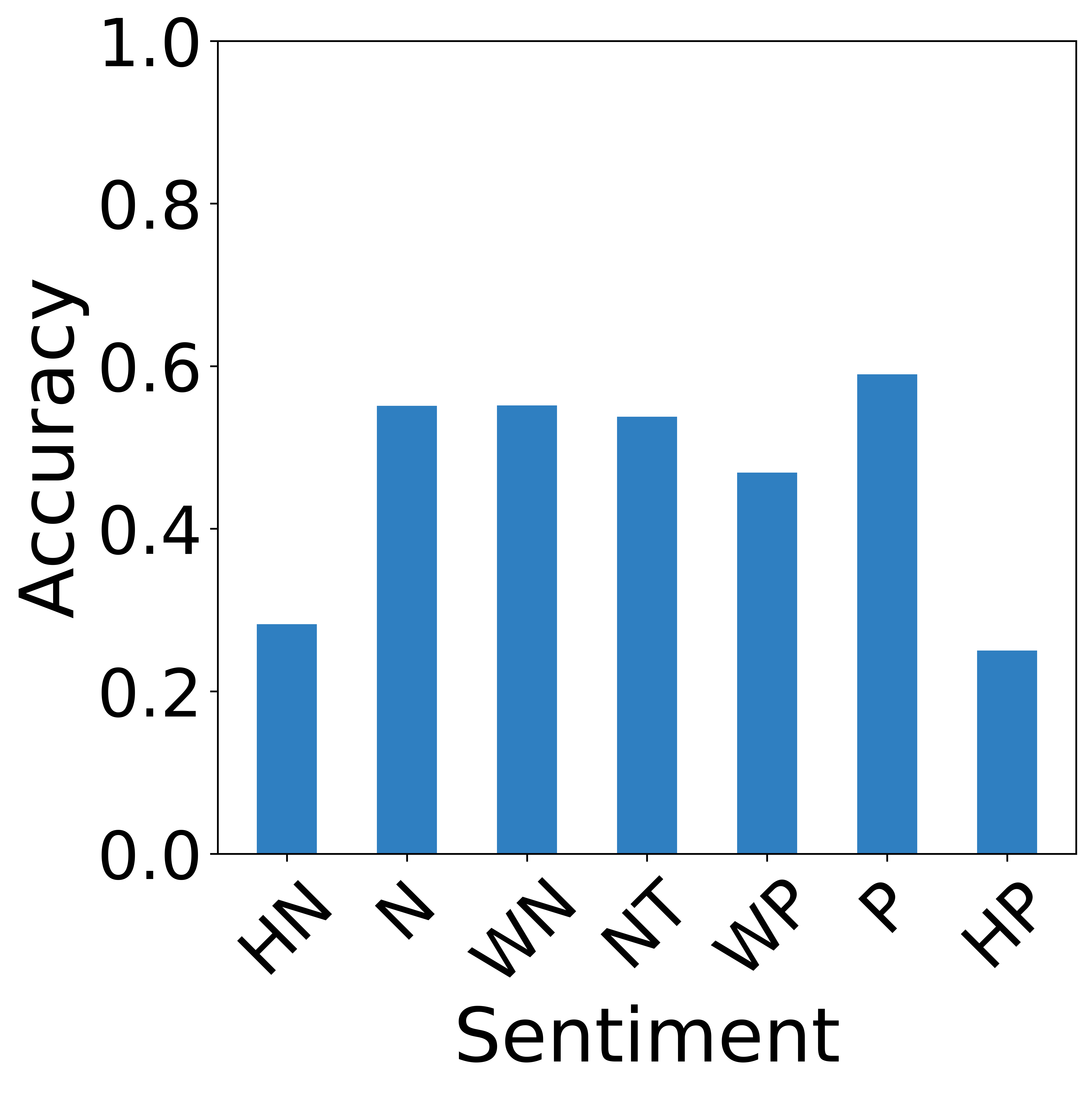}
    \label{fig:second}
}
\caption{Analysis on the MOSI dataset. (a) Confusion matrix, and (b) Corresponding accuracy for each sentiment.
HN: Highly Negative; N: Negative; WN: Weakly Negative; WP: Weak Positive; P: Positive; HP: Highly Positive.}
\label{fig:matrix}
\end{figure}

\begin{table}[t]
\small
\centering
\caption{Different masking signals.}
\setlength{\tabcolsep}{1mm}
\begin{tabular}{ccccc}
\toprule
Method & Acc-7($\uparrow$) & Acc-2($\uparrow$) & F1($\uparrow$)  & MAE($\downarrow$) \\
\midrule
\textbf{\papername (Ours)} & \textbf{51.46} & \textbf{90.09} & \textbf{90.08}  & \textbf{0.557} \\ 
Entropy    & 50.29  & 88.72 & 88.75  & 0.578 \\
Attention  & 50.48 & 88.63 & 88.62  & 0.572 \\
Random     & 49.27  & 88.33 & 88.31  & 0.575 \\
KL-div	& 50.32	& 89.30	& 89.20	& 0.612 \\
Cosine	& 47.96	& 88.90	& 88.71	& 0.628 \\
$L_2$	& 47.23	& 88.42	& 88.06	& 0.698 \\

\bottomrule
\end{tabular}
\label{Tab: masking}
\end{table}

\textbf{Masking Signals.}
To further validate the superiority of the proposed $I^2CS$-driven masking strategy, we conduct a comprehensive comparison against several common token-compression heuristics and distance-based criteria, as summarized in Table~\ref{Tab: masking}. 
The empirical results demonstrate that our \papername consistently achieves the best performance across all evaluation metrics. Notably, it reaches a peak of 51.46\% in the fine-grained Acc-7 metric and significantly reduces the MAE to 0.557, indicating superior prediction calibration and robustness. 
While traditional heuristics such as \textit{Entropy} and \textit{Attention} marginally outperform \textit{Random} masking, their reliance on signal uncertainty or saliency alone proves insufficient for distinguishing redundant noise from complex yet essential semantic cues. Furthermore, distance-based methods like \textit{Cosine} and $L_2$ similarity even underperform the random baseline, suggesting that proximity in the feature space does not inherently imply semantic redundancy and can even lead to the loss of discriminative structures. 
In contrast, our \papername shifts the pruning paradigm from simple saliency to semantic distribution consistency. By explicitly measuring the agreement within and across modalities, \papername effectively identifies and removes tokens that induce predictions inconsistent with the multimodal consensus. \textit{KL-div} is the most similar method to ours, but it still underperforms \papername, which indicates that JS score is a little better than KL divergence in this design. This consistency-aware criterion provides a more reliable importance signal, ensuring that the remaining tokens form a logically coherent and task-relevant representation, thereby making \papername an effective and practical choice for efficient multimodal learning.

\subsection{Visualization}
\textbf{Visualizing Token Consistency.} To better assess the effectiveness of \papername, we visualize token-level intra- and inter-modality semantic consistency in Fig.\,\ref{fig:token_heatmap}. Before optimization, tokens exhibit weak consistency both within and across modalities. After applying our consistency-aware training, consistency improves substantially in both aspects, explaining the robustness of \papername under aggressive token compression. These results confirm that \papername enhances representational quality by explicitly modeling and supervising semantic consistency during training.

 \textbf{Confusion Matrix.} To further investigate the effect of sentiment granularity on MSA, as shown in Fig.~\ref{fig:matrix}, we examine the confusion matrix and per-class accuracy on MOSI. Most sentiment categories achieve accuracies above 50\%; however, ``HN’’ and particularly ``HP’’ exhibit substantially lower performance, thereby constraining the overall 7-class accuracy. A closer inspection of the confusion matrix shows that ``HN’’ and ``HP’’ contain only 46 and 20 samples, respectively, underscoring the influence of the dataset’s long-tailed distribution. This imbalance poses a critical challenge for fine-grained sentiment understanding and motivates future exploration in this direction.

\section{Conclusion}
In this paper, we propose \papername, a consistency-aware framework for MSA that explicitly models semantic distribution consistency within and across modalities. By computing $I^2CS$ in a shared prediction space, \papername provides a unified signal for consistency regularization, relevance-aware reweighting, and token compression.
Extensive experiments on three benchmarks validate that \papername outperforms prior state-of-the-art approaches and remains robust under tight token budgets, maintaining competitive performance even after discarding 90\% tokens. Overall, our results suggest that explicitly optimizing semantic consistency is a practical and effective route to building multimodal sentiment models that are both accurate and compute-efficient.

\section*{Acknowledgements}
This work was primarily conducted during an internship at Amazon. Additional support was provided by the National Science Foundation (NSF) under grants CNS-2328972, CCF-2324937, and CNS-2122320, as well as the National Institutes of Health (NIH) under grant R01EB033387.

\section*{Impact Statement}
(i) This study reveals the importance of modeling intra- and inter-modality consistency, offering a new perspective for exploiting multimodal learning.
(ii) The proposed consistency-aware learning paradigm is general and flexible, allowing \papername to be readily extended to other multimodal tasks by adapting its scoring inputs. 
Future work will explore robust consistency learning under modality-missing settings.

\bibliography{refs}

@inproceedings{fang2025emoe,
  title={Emoe: Modality-specific enhanced dynamic emotion experts},
  author={Fang, Yiyang and Huang, Wenke and Wan, Guancheng and Su, Kehua and Ye, Mang},
  booktitle={Proceedings of the Computer Vision and Pattern Recognition Conference},
  pages={14314--14324},
  year={2025}
}

@inproceedings{yu2020ch,
  title={Ch-sims: A chinese multimodal sentiment analysis dataset with fine-grained annotation of modality},
  author={Yu, Wenmeng and Xu, Hua and Meng, Fanyang and Zhu, Yilin and Ma, Yixiao and Wu, Jiele and Zou, Jiyun and Yang, Kaicheng},
  booktitle={Proceedings of the 58th annual meeting of the association for computational linguistics},
  pages={3718--3727},
  year={2020}
}

@article{ZHU2025102787,
title = {Multimodal sentiment analysis with unimodal label generation and modality decomposition},
journal = {Information Fusion},
volume = {116},
pages = {102787},
year = {2025},
issn = {1566-2535},
author = {Linan Zhu and Hongyan Zhao and Zhechao Zhu and Chenwei Zhang and Xiangjie Kong},
}

@inproceedings{jian2026mdf,
  title={{MDF}: A Modality-Aware Disentanglement and Fusion Framework for Multimodal Sentiment Analysis},
  author={Jian, Zhongquan and Lv, Wenhan and Chen, Yanhao and Luo, Guanran and Qiu, Wentao and Wang, Shaopan and Hu, Bingbing and Wu, Qingqiang},
  booktitle={Proceedings of the AAAI Conference on Artificial Intelligence},
  volume={40},
  number={37},
  pages={31292--31300},
  year={2026}
}

@inproceedings{tong2024flashsloth,
  title={Flashsloth: Lightning multimodal large language models via embedded visual compression},
  author={Tong, Bo and Lai, Bokai and Zhou, Yiyi and Luo, Gen and Shen, Yunhang and Li, Ke and Sun, Xiaoshuai and Ji, Rongrong},
  booktitle={Proceedings of the Computer Vision and Pattern Recognition Conference},
  pages={14570--14581},
  year={2025}
}

@inproceedings{mao2022m,
  title={M-SENA: An integrated platform for multimodal sentiment analysis},
  author={Mao, Huisheng and Yuan, Ziqi and Xu, Hua and Yu, Wenmeng and Liu, Yihe and Gao, Kai},
  booktitle={Proceedings of the 60th Annual Meeting of the Association for Computational Linguistics: System Demonstrations},
  pages={204--213},
  year={2022}
}

@inproceedings{han2021improving,
  title={Improving multimodal fusion with hierarchical mutual information maximization for multimodal sentiment analysis},
  author={Han, Wei and Chen, Hui and Poria, Soujanya},
  booktitle={Proceedings of the 2021 conference on empirical methods in natural language processing},
  pages={9180--9192},
  year={2021}
}

@inproceedings{he2024zipvl,
  title={ZipVL: Accelerating Vision-Language Models through Dynamic Token Sparsity},
  author={He, Yefei and Chen, Feng and Liu, Jing and Shao, Wenqi and Zhou, Hong and Zhang, Kaipeng and Zhuang, Bohan},
  booktitle={Proceedings of the IEEE/CVF International Conference on Computer Vision},
  pages={20477--20486},
  year={2025}
}

@inproceedings{liu2018efficient,
  title={Efficient low-rank multimodal fusion with modality-specific factors},
  author={Liu, Zhun and Shen, Ying and Lakshminarasimhan, Varun Bharadhwaj and Liang, Paul Pu and Zadeh, AmirAli Bagher and Morency, Louis-Philippe},
  booktitle={Proceedings of the 56th Annual Meeting of the Association for Computational Linguistics (Volume 1: Long Papers)},
  pages={2247--2256},
  year={2018}
}

@article{wang2025models,
title={From Models to Systems: A Comprehensive Survey of Efficient Multimodal Learning},
author={Pan Wang and Siwei Song and Hui Ji and Siqi Cao and Heng Yu and Zhijian Liu and Huanrui Yang and Yingyan Celine Lin and Beidi Chen and Mohit Bansal and Xiaoming Liu and Pengfei Zhou and Ming-Hsuan Yang and Tianlong Chen and Jingtong Hu},
journal={Transactions on Machine Learning Research},
issn={2835-8856},
year={2026},
}

@inproceedings{devlin2019bert,
  title={Bert: Pre-training of deep bidirectional transformers for language understanding},
  author={Devlin, Jacob and Chang, Ming-Wei and Lee, Kenton and Toutanova, Kristina},
  booktitle={Proceedings of the 2019 conference of the North American chapter of the association for computational linguistics: human language technologies, volume 1 (long and short papers)},
  pages={4171--4186},
  year={2019}
}

@inproceedings{baevski2022data2vec,
  title={Data2vec: A general framework for self-supervised learning in speech, vision and language},
  author={Baevski, Alexei and Hsu, Wei-Ning and Xu, Qiantong and Babu, Arun and Gu, Jiatao and Auli, Michael},
  booktitle={International conference on machine learning},
  pages={1298--1312},
  year={2022},
  organization={PMLR}
}

@article{liu2026plug,
  title={Plug-and-play Feature Causality Decomposition for Multimodal Representation Learning},
  author={Liu, Ye and Ji, Zihan and Cai, Hongmin},
  journal={Advances in Neural Information Processing Systems},
  volume={38},
  pages={58912--58938},
  year={2026}
}

@article{liu2019roberta,
  title={Roberta: A robustly optimized bert pretraining approach},
  author={Liu, Yinhan and Ott, Myle and Goyal, Naman and Du, Jingfei and Joshi, Mandar and Chen, Danqi and Levy, Omer and Lewis, Mike and Zettlemoyer, Luke and Stoyanov, Veselin},
  journal={arXiv preprint arXiv:1907.11692},
  year={2019}
}

@inproceedings{fuglede2004jensen,
  title={Jensen-Shannon divergence and Hilbert space embedding},
  author={Fuglede, Bent and Topsoe, Flemming},
  booktitle={International symposium onInformation theory, 2004. ISIT 2004. Proceedings.},
  pages={31},
  year={2004},
  organization={IEEE}
}

@inproceedings{zhang2024pmod,
  title={p-MoD: Building mixture-of-depths mllms via progressive ratio decay},
  author={Zhang, Jun and Meng, Desen and Zhang, Zhengming and Huang, Zhenpeng and Wu, Tao and Wang, Limin},
  booktitle={Proceedings of the IEEE/CVF International Conference on Computer Vision},
  pages={3705--3715},
  year={2025}
}

@article{xing2025vision,
  title={Vision-centric token compression in large language model},
  author={Xing, Ling and Wang, Alex Jinpeng and Yan, Rui and Shu, Xiangbo and Tang, Jinhui},
  journal={Advances in Neural Information Processing Systems},
  volume={38},
  pages={33080--33110},
  year={2026}
}

@inproceedings{chen2024fastv,
  title={An image is worth 1/2 tokens after layer 2: Plug-and-play inference acceleration for large vision-language models},
  author={Chen, Liang and Zhao, Haozhe and Liu, Tianyu and Bai, Shuai and Lin, Junyang and Zhou, Chang and Chang, Baobao},
  booktitle={European Conference on Computer Vision},
  pages={19--35},
  year={2024},
  organization={Springer}
}

@inproceedings{yang2024clgsi,
  title={CLGSI: a multimodal sentiment analysis framework based on contrastive learning guided by sentiment intensity},
  author={Yang, Yang and Dong, Xunde and Qiang, Yupeng},
  booktitle={Findings of the Association for Computational Linguistics: NAACL 2024},
  pages={2099--2110},
  year={2024}
}

@inproceedings{wu2024multimodal,
  title={Multimodal Multi-loss Fusion Network for Sentiment Analysis},
  author={Wu, Zehui and Gong, Ziwei and Koo, Jaywon and Hirschberg, Julia},
  booktitle={Proceedings of the 2024 Conference of the North American Chapter of the Association for Computational Linguistics: Human Language Technologies (Volume 1: Long Papers)},
  pages={3588--3602},
  year={2024}
}

@inproceedings{wang2025dlf,
  title={{DLF}: Disentangled-language-focused multimodal sentiment analysis},
  author={Wang, Pan and Zhou, Qiang and Wu, Yawen and Chen, Tianlong and Hu, Jingtong},
  booktitle={Proceedings of the AAAI Conference on Artificial Intelligence},
  pages={21180--21188},
  year={2025}
}

@inproceedings{yang2022learning,
  title={Learning modality-specific and-agnostic representations for asynchronous multimodal language sequences},
  author={Yang, Dingkang and Kuang, Haopeng and Huang, Shuai and Zhang, Lihua},
  booktitle={Proceedings of the 30th ACM International Conference on Multimedia},
  pages={1708--1717},
  year={2022}
}

@inproceedings{tsai2018learning,
  title={Learning factorized multimodal representations},
  author={Tsai, Yao-Hung Hubert and Liang, Paul Pu and Zadeh, Amir and Morency, Louis-Philippe and Salakhutdinov, Ruslan},
  booktitle={International Conference on Learning Representations},
  year={2019},
}

@inproceedings{yang2022disentangled,
  title={Disentangled representation learning for multimodal emotion recognition},
  author={Yang, Dingkang and Huang, Shuai and Kuang, Haopeng and Du, Yangtao and Zhang, Lihua},
  booktitle={Proceedings of the 30th ACM International Conference on Multimedia},
  pages={1642--1651},
  year={2022}
}

@inproceedings{williams2018dnn,
  title={Dnn multimodal fusion techniques for predicting video sentiment},
  author={Williams, Jennifer and Comanescu, Ramona and Radu, Oana and Tian, Leimin},
  booktitle={Proceedings of grand challenge and workshop on human multimodal language (Challenge-HML)},
  pages={64--72},
  year={2018}
}

@inproceedings{williams2018recognizing,
  title={Recognizing emotions in video using multimodal dnn feature fusion},
  author={Williams, Jennifer and Kleinegesse, Steven and Comanescu, Ramona and Radu, Oana},
  booktitle={Proceedings of Grand Challenge and Workshop on Human Multimodal Language (Challenge-HML)},
  pages={11--19},
  year={2018}
}

@article{geetha2024multimodal,
  title={Multimodal emotion recognition with deep learning: advancements, challenges, and future directions},
  author={Geetha, A.V. and Mala, T. and Priyanka, D and Uma, EJIF},
  journal={Information Fusion},
  volume={105},
  pages={102218},
  year={2024},
  publisher={Elsevier}
}

@article{zadeh2016multimodal,
  title={Multimodal sentiment intensity analysis in videos: Facial gestures and verbal messages},
  author={Zadeh, Amir and Zellers, Rowan and Pincus, Eli and Morency, Louis-Philippe},
  journal={IEEE Intelligent Systems},
  volume={31},
  number={6},
  pages={82--88},
  year={2016},
  publisher={IEEE}
}

@inproceedings{zadeh2017tensor,
  title={Tensor fusion network for multimodal sentiment analysis},
  author={Zadeh, Amir and Chen, Minghai and Poria, Soujanya and Cambria, Erik and Morency, Louis-Philippe},
  booktitle={Proceedings of the 2017 conference on empirical methods in natural language processing},
  pages={1103--1114},
  year={2017}
}

@inproceedings{hazarika2020misa,
  title={Misa: Modality-invariant and-specific representations for multimodal sentiment analysis},
  author={Hazarika, Devamanyu and Zimmermann, Roger and Poria, Soujanya},
  booktitle={Proceedings of the 28th ACM international conference on multimedia},
  pages={1122--1131},
  year={2020}
}

@inproceedings{lv2021progressive,
  title={Progressive modality reinforcement for human multimodal emotion recognition from unaligned multimodal sequences},
  author={Lv, Fengmao and Chen, Xiang and Huang, Yanyong and Duan, Lixin and Lin, Guosheng},
  booktitle={Proceedings of the IEEE/CVF Conference on Computer Vision and Pattern Recognition},
  pages={2554--2562},
  year={2021}
}

@inproceedings{liang2021attention,
  title={Attention is not enough: Mitigating the distribution discrepancy in asynchronous multimodal sequence fusion},
  author={Liang, Tao and Lin, Guosheng and Feng, Lei and Zhang, Yan and Lv, Fengmao},
  booktitle={Proceedings of the IEEE/CVF International Conference on Computer Vision},
  pages={8148--8156},
  year={2021}
}

@inproceedings{zadeh2018multimodal,
  title={Multimodal language analysis in the wild: Cmu-mosei dataset and interpretable dynamic fusion graph},
  author={Zadeh, AmirAli Bagher and Liang, Paul Pu and Poria, Soujanya and Cambria, Erik and Morency, Louis-Philippe},
  booktitle={Proceedings of the 56th Annual Meeting of the Association for Computational Linguistics},
  pages={2236--2246},
  year={2018}
}

@inproceedings{zadeh2018memory,
  title={Memory fusion network for multi-view sequential learning},
  author={Zadeh, Amir and Liang, Paul Pu and Mazumder, Navonil and Poria, Soujanya and Cambria, Erik and Morency, Louis-Philippe},
  booktitle={Proceedings of the AAAI conference on artificial intelligence},
  pages={5634--5641},
  year={2018}
}

@inproceedings{tsai2019multimodal,
  title={Multimodal transformer for unaligned multimodal language sequences},
  author={Tsai, Yao-Hung Hubert and Bai, Shaojie and Liang, Paul Pu and Kolter, J Zico and Morency, Louis-Philippe and Salakhutdinov, Ruslan},
  booktitle={Proceedings of the conference. Association for Computational Linguistics. Meeting},
  volume={2019},
  pages={6558},
  year={2019},
  organization={NIH Public Access}
}

@inproceedings{yu2021learning,
  title={Learning modality-specific representations with self-supervised multi-task learning for multimodal sentiment analysis},
  author={Yu, Wenmeng and Xu, Hua and Yuan, Ziqi and Wu, Jiele},
  booktitle={Proceedings of the AAAI conference on artificial intelligence},
  pages={10790--10797},
  year={2021}
}

@inproceedings{li2023decoupled,
  title={Decoupled Multimodal Distilling for Emotion Recognition},
  author={Li, Yong and Wang, Yuanzhi and Cui, Zhen},
  booktitle={Proceedings of the IEEE/CVF Conference on Computer Vision and Pattern Recognition},
  pages={6631--6640},
  year={2023}
}

@inproceedings{rahman2020integrating,
  title={Integrating multimodal information in large pretrained transformers},
  author={Rahman, Wasifur and Hasan, Md Kamrul and Lee, Sangwu and Zadeh, Amir and Mao, Chengfeng and Morency, Louis-Philippe and Hoque, Ehsan},
  booktitle={Proceedings of the conference. Association for Computational Linguistics. Meeting},
  volume={2020},
  pages={2359},
  year={2020},
  organization={NIH Public Access}
}
\bibliographystyle{icml2026}


\newpage
\appendix
\section{Appendix}

\subsection{Configuration of Hyperparameters}
\label{sec:setup}

Table~\ref{tab:hypa} presents the configuration of all hyperparameters of the proposed \papername for each benchmark. These hyperparameters are optimized based on performance evaluations conducted on the validation dataset. Training a model takes from 30 minutes to several hours on a single NVIDIA A100-40G GPU.

\begin{table}[h]
  \centering
  \small
  \setlength\tabcolsep{1.0mm}
  \caption{Hyperparameter settings on all MSA benchmarks. T: Text, A: Audio. The token length is based on the BERT/RoBERTa-based backbone.}
  \begin{tabular}{lccc}
    \toprule
    Setting & MOSI 
    &MOSEI & CH-SIMS\\
    \midrule
    Optimizer &AdamW &AdamW & AdamW\\
    Learning rate &5e-6&5e-6 & 1e-5\\
    Batch size&8& 8 & 8\\
    Token length  &50/96&50/96&50/96\\
    Early stopping epochs &8 &8 &8\\
    $\alpha$   &1.0&1.0 & 1.0\\
    $\beta$   &1.0&1.0  & 1.0\\
    $\gamma$ & 20.0 & 20.0 & 20.0 \\
    $\lambda_a$  & 1.0 & 1.0 & 1.0 \\
    $\lambda_t$  & 1.0 & 1.0 & 1.0 \\
    $[\lambda_1, \lambda_2,\lambda_3]$ & [1, 1, 1] & [1, 0.3, 1] & [1, 1, 1] \\
    \bottomrule
  \end{tabular}
  \label{tab:hypa}
\end{table}

\subsection{Details for the Three-Modality Implementation}
As shown in Tables~\ref{tab:comparison}, \ref{tab:tav_ta}, and \ref{tab:ch_sims}, to ensure fairness, when extending our approach to the three-modality case, we compute the proposed score for each modality by considering its interaction with the other two modalities. Specifically, for a given target modality, we first calculate the score between the target modality and each of the remaining two modalities independently. The final score for this modality is then obtained by averaging these pairwise scores. This averaged score is subsequently used to guide the token optimization process of the target modality. In this way, our method can be naturally generalized from the bimodal setting to the trimodal setting, while maintaining consistent optimization principles.

\subsection{Additional Results}

\textbf{Impact of Different Modality in the RoBERTa-based Setting (T+A).}
Table~\ref{tab:at} presents a comparative analysis between our multimodal \papername framework and its unimodal counterparts on the MOSI benchmark. The results demonstrate a clear superiority of multimodal integration; \papername achieves an Acc-7 of 51.46\%, outperforming the ``only Text" and ``only Audio" baselines by 5.71\% and 12.35\%, respectively. This substantial margin underscores the framework's capability to effectively capture and fuse complementary information across disparate modalities. Furthermore, while the ``only Text" setting yields higher accuracy than ``only Audio", the fused model's dominance confirms that \papername successfully leverages acoustic features as essential contextual signals to resolve unimodal ambiguities.

\textbf{Impact of Modality Pairs in the BERT-based Setting (T+A+V).}
To investigate the contribution of specific modality pairs to the final consistency signal, we conduct an ablation study in the BERT-based setting (Table~\ref{tab:avt}). The ``All" configuration, incorporating consistency signals from all modality pairs, achieves the highest performance across all metrics (e.g., 47.96\% Acc-7 and 0.698 MAE). Removing any modality pair leads to a performance degradation, validating the necessity of a holistic cross-modal consensus. Notably, pairs involving the text modality, specifically (T, A) and (V, T), provide more significant individual contributions. This suggests that linguistic cues serve as a more reliable anchor for establishing decision-level agreement in sentiment analysis. The cumulative benefit observed in the ``All" setting indicates that \papername benefits from a convergent validation process across all available modality perspectives.

\begin{table}[t]
\centering
\caption{Performance of different modalities on the MOSI benchmark in the setting of A+T. A: Audio, T: Text.}
\setlength{\tabcolsep}{0.2mm}
\begin{tabular}{lcccc}
\toprule
\textbf{Modality} & \textbf{Acc-7 (\%)} & \textbf{Acc-2 (\%)} & \textbf{F1 (\%)} & \textbf{MAE($\downarrow$)} \\
\midrule
\textbf{\papername (Ours)} & \textbf{51.46} & \textbf{90.09} & \textbf{90.08} & \textbf{0.557} \\
\midrule
only Audio & 39.11 & 84.23 & 84.25 & 0.805 \\
only Text & 45.75 & 86.61 & 86.62 & 0.654 \\
\bottomrule
\end{tabular}
\label{tab:at}
\end{table}

\begin{table}[t]
\centering
\caption{Ablation of I²CS pair on the MOSI dataset in the BERT-based setting. T: Text, A: Audio, V: Visual.}
\setlength{\tabcolsep}{0.6mm}
\begin{tabular}{lcccc}
\toprule
\textbf{Modality Pair} &	\textbf{Acc-7 (\%)} &	\textbf{Acc-2 (\%)} & \textbf{F1 (\%)}	& \textbf{MAE($\downarrow$)} \\
\midrule
\textbf{All (Ours)}	& \textbf{47.96}	& \textbf{86.06}	& \textbf{86.06}	& \textbf{0.698} \\
\midrule
(A, V)	& 43.44	& 83.58	& 83.02	& 0.771 \\
(V, T)	& 45.92	& 84.49	& 84.50	& 0.732 \\
(T, A)	& 47.23	& 85.26	& 85.21	& 0.728 \\
\bottomrule
\end{tabular}
\label{tab:avt}
\end{table}

\begin{table*}[t]
\small
\centering
\caption{Performance of RoBERTa-based ConsMSA with ``T+A+V'' and ``T+A'' on the MOSI dataset. T: Text, A: Audio, V: Visual.}
\setlength{\tabcolsep}{0.8mm}
\begin{tabular}{l c cccc cccc}
\toprule
\multirow{2}{*}{Method} & \multirow{2}{*}{Modality} & \multicolumn{4}{c}{CMU-MOSI} & \multicolumn{4}{c}{CMU-MOSEI} \\
\cmidrule(lr){3-6} \cmidrule(lr){7-10}
& & Acc-7~$\uparrow$ & Acc-2~$\uparrow$ & F1~$\uparrow$ & MAE~$\downarrow$ & Acc-7~$\uparrow$ & Acc-2~$\uparrow$ & F1~$\uparrow$ & MAE~$\downarrow$ \\
\midrule
MMML                 & T+A   & 50.34 & 89.69 & 89.67 & 0.580 & 55.74 & 88.02 & 88.15 & 0.492 \\
\textbf{\papername (Ours)} & T+A   & \textbf{51.46} & \textbf{90.09} & \textbf{90.08} & \textbf{0.557} & \textbf{56.00} & \textbf{88.41} & \textbf{88.39} & \textbf{0.485} \\
FCD                 & T+A+V   & - & 88.48 & 88.44 & 0.599 & - & 86.63 & 86.63 & 0.500 \\
\textbf{\papername (Ours)} & T+A+V   & \textbf{51.60} & \textbf{90.48} & \textbf{90.38} & \textbf{0.544} & \textbf{56.08} & \textbf{88.89} & \textbf{88.80} & \textbf{0.483} \\
\bottomrule
\end{tabular}
\label{tab:tav_ta}
\end{table*}

\textbf{Impact of Modality Selection in the RoBERTa-based Setting.} 
To ensure a fair comparison with state-of-the-art (SOTA) models, our main results focus on the bi-modal (``T+A'') configuration. However, to comprehensively analyze the impact of the visual modality, we evaluate both the bi-modal and tri-modal (``T+A+V'') settings in Table~\ref{tab:tav_ta}. The results consistently demonstrate the robustness of ConsMSA across both configurations. In the absence of visual inputs (T+A), ConsMSA consistently outperforms the strong MMML baseline; on CMU-MOSI, it achieves an Acc-2 of 90.09\% and reduces the MAE to 0.557, indicating highly resilient semantic extraction from limited modalities. Furthermore, when provided with complete tri-modal inputs (T+A+V), ConsMSA significantly surpasses the FCD~\citep{liu2026plug} baseline, yielding a remarkable Acc-2 of 90.48\% and an F1 score of 90.38\% on CMU-MOSI, thereby validating our architecture's advantage in multimodal alignment. Crucially, comparing performance across these settings reveals that integrating the visual modality yields steady enhancements across all metrics on both datasets---such as the F1 score rising from 88.39\% to 88.80\% and the MAE dropping from 0.485 to 0.483 on CMU-MOSEI. This confirms that ConsMSA effectively leverages complementary cross-modal information while successfully mitigating the risk of negative transfer from potential modality noise or redundancy.

\begin{table}[t]
\small
\centering
\caption{Comparison on CH-SIMS dataset.}
\label{tab:ch_sims}
\setlength{\tabcolsep}{0.8mm}
\begin{tabular}{llccccc}
\toprule
\textbf{Method} & \textbf{Modality} & \textbf{Acc2}~$\uparrow$ & \textbf{Acc3}~$\uparrow$ & \textbf{Acc5}~$\uparrow$ & \textbf{F1}~$\uparrow$ & \textbf{MAE}~$\downarrow$ \\
\midrule
MMML  & T+A   & 82.93 & 69.37 & 49.38 & 82.90 & 0.332 \\
\midrule
\textbf{Ours}   & \textbf{T+A}   & \textbf{83.59} & \textbf{70.68} & \textbf{50.77} & \textbf{83.64} & \textbf{0.331} \\
\textbf{Ours}   & \textbf{T+A+V} & \textbf{83.81} & \textbf{71.12} & \textbf{50.92} & \textbf{83.81} & \textbf{0.316} \\
\bottomrule
\end{tabular}
\end{table}

\textbf{Generalization on the CH-SIMS Dataset.} 
To further evaluate the generalization capability of our approach across different linguistic contexts, we conduct comparative experiments on the Chinese MSA dataset, CH-SIMS. As shown in Table~\ref{tab:ch_sims}, our method consistently delivers superior performance. Under the bi-modal (T+A) setting, our approach effectively outperforms the MMML baseline across all evaluation metrics. Most notably, it yields a 1.31\% absolute improvement in Acc-3 (from 69.37\% to 70.68\%) and a 1.39\% improvement in Acc-5 (from 49.38\% to 50.77\%), highlighting our model's capability to learn highly discriminative and robust multimodal representations even with restricted modality inputs. Furthermore, when extending to the full tri-modal (T+A+V) configuration, our method achieves peak performance, reaching 83.81\% for both Acc-2 and F1 score, while further lowering the MAE to 0.316. The consistent empirical gains observed when scaling from T+A to T+A+V confirm that our architecture effectively integrates supplementary visual signals without suffering from negative transfer, thereby reinforcing the universality and robustness of our end-to-end multimodal semantic consistency strategy.

\textbf{Hyperparameter Sensitivity.}
To investigate the robustness of our framework against varying penalty weights, we conduct a sensitivity analysis on the hyperparameters $\alpha$, $\beta$, and $\gamma$ on the CH-SIMS dataset. As summarized in Table~\ref{tab:hyperparameter_sensitivity}, we sweep the configuration space with $\alpha, \beta \in \{0.5, 1.0\}$ and $\gamma \in \{10, 20\}$. The empirical results exhibit remarkable stability across different settings. Most notably, core classification metrics such as Acc-2 (83.59\%) and Acc-3 (70.68\%) remain strictly identical despite the parameter variations, while the F1 score and MAE exhibit only negligible fluctuations. We attribute this strong resilience to the design of our consistency-based token selection mechanism. Specifically, as long as the relative importance of the semantic distribution consistency is mathematically preserved, the soft gating effectively functions as a stable anchor, consistently focusing on the same high-confidence multimodal cues regardless of the absolute magnitude of the loss weights. This inherent architectural robustness significantly alleviates the burden of exhaustive hyperparameter tuning, making our approach practical and generalizable for diverse multimodal tasks.

\begin{table}[t]
\centering
\small
\caption{Hyperparameter sensitivity on the CH-SIMS dataset.}
\label{tab:hyperparameter_sensitivity}
\setlength{\tabcolsep}{1mm}
\begin{tabular}{cccccccc}
\toprule
$\alpha$ & $\beta$ & $\gamma$ & \textbf{Acc2}~$\uparrow$ & \textbf{Acc3}~$\uparrow$ & \textbf{Acc5}~$\uparrow$ & \textbf{F1}~$\uparrow$ & \textbf{MAE}~$\downarrow$ \\
\midrule
\textbf{1.0} & \textbf{1.0} & \textbf{20} & \textbf{83.59} & \textbf{70.68} & \textbf{50.77} & 83.64 & \textbf{0.331} \\
1.0 & 0.5 & 20 & 83.59 & 70.68 & 49.67 & \textbf{83.70} & 0.339 \\
0.5 & 1.0 & 10 & 83.59 & 70.68 & 49.67 & \textbf{83.70} & 0.339 \\
0.5 & 0.5 & 10 & 83.59 & 70.68 & 49.67 & \textbf{83.70} & 0.339 \\
\bottomrule
\end{tabular}
\end{table}

\textbf{Efficiency Analysis.} To evaluate the computational efficiency of our approach, we benchmark the inference latency and runtime memory on the MOSI dataset using an NVIDIA V100 GPU, as summarized in Table~\ref{tab:latency}. Our method demonstrates remarkable acceleration capabilities across both BERT- and RoBERTa-based backbones. Notably, under the high token compression setting (ratio of 0.1), our model achieves a significant 2.03$\times$ speedup (lowering latency to 11.04 ms) in the BERT-based configuration and a 2.16$\times$ speedup (42.84 ms) in the RoBERTa-based configuration compared to their respective baselines. To thoroughly investigate the architectural overhead, we profile the latency of our $\text{I}^2\text{CS}$ computation, which takes only about 2.2081 ms on the V100 GPU. This empirical result strongly demonstrates that the massive efficiency gains derived from Token Compression far outweigh the marginal computational cost of the Consistency Scorer. Furthermore, this substantial reduction in latency is achieved without introducing any additional memory burden; for instance, the runtime memory in the RoBERTa setting remains nearly identical to the baseline (1.3076 GB vs. 1.3069 GB). Overall, these findings highlight the practical superiority of our method, offering a highly efficient multimodal learning, especially for MSA.

\begin{table}[t]
\centering
\small
\caption{Comparison of efficiency on the MOSI dataset. 1.0 and 0.1 mean the token retention ratios.}
\label{tab:efficiency_comparison}
\setlength{\tabcolsep}{0.6mm}
\begin{tabular}{lccc}
\toprule
\textbf{Methods}  & \textbf{Latency (ms)} & \textbf{Runtime Memory (GB)} & \textbf{Speedup} \\
\midrule
\multicolumn{4}{c}{BERT-based} \\
\midrule
TFN                & 22.3689 & 0.0101 & \textbf{1.00x} \\
MULT               & 22.5551 & 0.0110 & 0.99x \\
MISA               & 33.9139 & 0.0254 & 0.66x \\
MAG-BERT           & 33.5336 & 0.0555 & 0.67x \\
MMIM               & 78.2354 & 0.0312 & 0.29x \\
DLF                & 65.2429 & 0.0275 & 0.34x \\
EMOE               & 37.1566 & 0.0257 & 0.60x \\
\textbf{Ours (1.0)}& 31.4535 & \textbf{0.0215} & 0.71x \\
\textbf{Ours (0.1)}& \textbf{11.0351} & \textbf{0.0214} & \textbf{2.03x} \\
\midrule
\multicolumn{4}{c}{RoBERTa-based} \\
\midrule
MMML & 92.5583 & \textbf{1.3069} & 1.00x \\
\textbf{Ours (1.0)}& 94.9552 & \textbf{1.3073} & 0.97x \\
\textbf{Ours (0.1)}& \textbf{42.8431} & \textbf{1.3076} & \textbf{2.16x} \\
\bottomrule
\end{tabular}
\label{tab:latency}
\end{table}

\subsection{Brief Introduction of All Baselines}
We briefly introduce all baseline methods discussed in the main paper:
\begin{itemize}
    \item \textbf{EF-LSTM}: An early fusion model that concatenates multimodal features and feeds them into a unified LSTM for sentiment prediction.

    \item \textbf{LF-DNN}: This model employs an intermediate fusion approach, combining modality-specific weights during the learning phase.

    \item \textbf{TFN}: It captures intra-modal patterns using three dedicated subnetworks and models inter-modal interactions through unimodal, bimodal, and trimodal subtensors derived via a tensor fusion strategy.

    \item \textbf{LMF}: Utilizes a low-rank tensor approximation to efficiently fuse high-dimensional multimodal data, achieving scalability with the number of modalities.

    \item \textbf{MFN}: Proposes a memory fusion mechanism that performs both view-specific and cross-view processing over sequential inputs, generating predictions from the combined output.

    \item \textbf{Graph-MFN}: Extends MFN by introducing a Dynamic Fusion Graph to model cross-modal interactions as a graph structure, replacing the original Delta-memory Attention Network.

    \item \textbf{MulT}: Applies Transformer layers for soft cross-modal alignment, extending temporal receptive fields and introducing a ternary-symmetric architecture for feature fusion.

    \item \textbf{PMR}: Incorporates a progressive reinforcement process via a message hub and cyclic cross-modal attention, allowing gradual feature refinement among modalities.

    \item \textbf{MISA}: Inspired by domain separation frameworks, it decomposes each modality into modality-shared and modality-private spaces, facilitating structured fusion.

    \item \textbf{MAG-BERT}: Introduces a Multimodal Adaptation Gate that enables BERT to absorb nonverbal cues during the fine-tuning process.

    \item \textbf{FDMER}: Proposes FDMER framework, which disentangles features into modality-invariant (common) and modality-specific (private) subspaces via dedicated encoders and adversarial training, encouraging both shared understanding and modality uniqueness.

    \item \textbf{DMD}: Implements a learnable Graph Unit to dynamically guide knowledge distillation between modalities, enabling adaptable and resilient representation learning.

    \item \textbf{MMIM}: Proposes MultiModal InfoMax (MMIM), a novel framework for multimodal sentiment analysis (MSA) that explicitly maximizes mutual information to preserve task-relevant information during fusion.

    \item \textbf{Self-MM}: Replaces hand-labeled unimodal annotations using a self-supervised label generation module. It jointly trains multimodal and unimodal tasks to learn both consistency and difference across modalities, using dynamic weighting to focus on challenging samples.

    \item \textbf{DLF}: A Disentangled-Language-Focused framework for MSA that explicitly separates shared and modality-specific features, enhances language representations via a language-guided attractor, and adopts hierarchical prediction to improve accuracy.

    \item \textbf{ULMD}: Proposes a decomposed multi-task learning framework based on unimodal label generation and modality decomposition (ULMD), which consists of one multimodal task and three unimodal tasks.

    \item \textbf{EMOE}: Introduces an adaptive multimodal fusion approach that combines a Mixture of Modality Experts with unimodal distillation to effectively address sample-varying modality importance while preserving modality-specific features, demonstrating broad effectiveness on diverse multimodal emotion recognition and multimodal intent recognition benchmarks.

    \item \textbf{MDF}: Proposes a Modality-aware Disentangle and Fusion (MDF) framework that establishes cross-modal heterogeneity through text-anchored disentanglement and enhancement, followed by a dynamically adaptive weighting mechanism for effective MSA.

    \item \textbf{FCD}: Proposes a causality-inspired decomposition strategy that applies backdoor adjustments to eliminate aleatoric uncertainty without altering the underlying fusion architecture.

    \item \textbf{MMML}: Proposes an optimized MSA framework, which integrates carefully selected feature encoders and fusion strategies across modalities. It further introduces multi-loss training and context integration, which significantly enhance representation learning.
    
\end{itemize}

\end{document}